\documentclass{article}

\PassOptionsToPackage{round}{natbib}

\usepackage[dblblindworkshop, final]{neurips_2026}

\usepackage[utf8]{inputenc} 
\usepackage[T1]{fontenc}    
\usepackage{hyperref}       
\usepackage{url}            
\usepackage{booktabs}       
\usepackage{array}          
\usepackage{caption}        
\usepackage{multirow}       
\usepackage{float}          
\usepackage{graphicx}       
\usepackage{amsfonts}       
\usepackage{nicefrac}       
\usepackage{microtype}      
\usepackage{xcolor}         

\title{
  IND-Eval: Aligning Evaluation and Simulation for Industrial Time Series Foundation Models
  }

\workshoptitle{Foundation Models for Temporal Systems}

\author{%
  Vebjørn Haugland\thanks{Corresponding author:
    \texttt{vebjorn.haugland@cognite.com}}
  \quad
  Pavel Gulin
  \quad
  David Foss
  \quad
  Peet Cremer
  \\
  \bfseries
  Erlend Aune
  \quad
  Simone Casolo
  \quad
  Andreas Solbrå
  \quad
  Milad Mobarhan
  \\[0.4em]
  \mdseries
  Cognite
}

\begin{document}

\maketitle
\setcounter{footnote}{0}

\begin{abstract}

  We introduce IND-Eval, a public benchmark that organizes industrial
  time series across domains into source-specific forecasting tasks. Relative to
  established benchmarks such as GIFT-Eval, it emphasizes finer sampling
  resolutions and a greater prevalence of multivariate forecasting. Evaluating
  Chronos-2, TiRex-2, and TimesFM-3, we find that TimesFM-3 achieves the lowest
  source-balanced aggregate nMASE and nCRPS and leads seven of nine sources
  under each metric; source-specific exceptions reinforce the need for
  domain-representative evaluation for model selection. These findings also
  motivate improving domain alignment during pretraining, where real industrial
  data are often too scarce to close domain-specific gaps directly; we therefore
  investigate whether a domain's statistical structure can guide simulator
  design. In a case study on two offshore well-monitoring regimes, we improve a
  general simulator's coverage using an oscillator-augmented simulation
  pipeline. This lowers forecasting error in the target domain but reduces performance
  on the subset of GIFT-Eval with natively univariate targets. A 50/50 mixture
  of the general and augmented simulation recipes achieves the
  lowest error among the three training recipes for all four event--metric pairs
  in the case study while recovering most of the lost GIFT-Eval performance.
  These results show how domain-representative evaluation can
  guide model selection and simulator composition while revealing a
  trade-off between specialization and generality.
\end{abstract}

\section{Introduction}
\label{sec:introduction}

Industrial time series support high-value decisions in factories,
energy systems, and other complex physical processes. Accurate
forecasts can help reduce operational risk, operating costs, and unplanned
downtime. Public industrial datasets exist, but evaluation remains fragmented
across domains, tasks, and use cases, making deployment-oriented model selection
difficult.
General leaderboards summarize
performance across heterogeneous tasks. An industrial deployment instead
concerns a particular process and operating regime, with specific target
variables, sampling rate, forecast horizon, and episode boundaries. Aggregate
scores can therefore mislead model selection for a process, asset, or
manufacturing batch, with potential safety consequences.

We therefore introduce IND-Eval,\footnote{We release IND-Eval on Hugging Face
at \url{https://huggingface.co/datasets/cognitedata/IND-eval}.} a public
benchmark comprising 187 source-defined forecasting tasks from nine public
sources across several
physical industrial sectors. It makes target variables, sampling rates,
forecast horizons, and episode boundaries explicit in evaluation.

IND-Eval complements broad forecasting
benchmarks~\citep{aksu2024gifteval,qiao2026time,shchur2025fev} and industrial
telemetry resources~\citep{cohen2025this,othman2026factorynet} by emphasizing
operational sources and underrepresented task structures.

This scope also changes how results should be interpreted. In our illustrative
comparison of Chronos-2~\citep{ansari2025chronos2},
TiRex-2~\citep{podest2026tirex2}, and
TimesFM-3~\citep{jain2026timesfm3}, performance varies by source, so aggregate
scores do not identify the best model for every setting. Results should instead
be interpreted within the task taxonomy represented by the evaluation
data~\citep{saqur2026seekingsota}. This intended-use evidence is also relevant
to safety and governance: for high-risk AI systems, the EU AI Act requires
testing against metrics appropriate to the intended purpose before market
placement or deployment~\citep{euaiact2024}. While IND-Eval does not establish
regulatory compliance, it can support the intended-purpose testing required
for high-risk AI systems under the EU AI Act.

Evaluation can also reveal how training data fail to represent a target use
case. TSFMs are increasingly trained on simulated
data~\citep{das2024timesfm,ansari2024chronos}, and simulator composition
materially affects downstream performance~\citep{nagpal2026mixdontpick}.
Where target data are available, fine-tuning can specialize a
model~\citep{qiao2025multiscale}; where target data are scarce, targeted
simulation is a complementary route.
We study this approach in an exploratory case study on severe slugging and flow
instability from 3W~\citep{vargas2026threew,vargas2019realistic}, a public
multivariate offshore well-monitoring dataset. Selected per-channel temporal
statistics show that SarSim0, a generic SARIMA-based simulation
pipeline~\citep{oreshkin2026zero}, poorly matches these regimes' smoothness,
spectral entropy, and temporal persistence. We therefore construct
SarSim0-oscillator, an augmented pipeline in which SarSim0-generated forcing
drives a coupled-oscillator network, and assess whether its generated windows
better match the target statistics and yield lower event-specific forecasting
error.

Our contributions are (1)~IND-Eval, 187 source-defined tasks exposing
underrepresented industrial structures and source-dependent model rankings;
and (2)~an exploratory 3W study linking target-statistic coverage, simulator
composition, and downstream forecasting.

\section{IND-Eval}
\label{sec:indeval}

\paragraph{Datasets.}\label{subsec:datasets}
IND-Eval comprises 187 forecasting tasks from nine public datasets: a
six-source \emph{industrial core} emphasizing task structures under-represented
in established forecasting benchmarks, and three industrially relevant sources
shared with GIFT-Eval.
NEATS~\citep{nopta_neats_2026} contains offshore drilling streams;
3W groups multivariate oil-well instances into ten operating-event classes
(normal plus nine undesirable events);
PRONTO~\citep{stief2019heterogeneous} contains six annotated multiphase-flow
transients with three known-future control inputs; Mesa del
Sol~\citep{bashir2023power} records 15 months of microgrid operation at three
resolutions; Tablet compression~\citep{vzagar2022big} covers five process
variables from pharmaceutical batches; and MetroPT-3~\citep{veloso2022metropt}
follows five channels of a metro-train air-production unit. The shared sources
ETT~\citep{zhou2021informer} (two seven-channel transformer series),
Electricity~\citep{trindade2015electricity} (370 load series), and
Solar~\citep{lai2018modeling} (137 simulated plants) use the same variants as
GIFT-Eval.

A task pairs one source-specific series group with one forecast horizon and
scores every target variate in each context--target window. Windows remain
within source instances, while known-future covariates and within-instance
operating transitions are retained where available
(Appendix~\ref{app:ind-composition}).

\begin{figure*}[t]
    \centering
    \makebox[\textwidth][c]{%
        \raisebox{-0.5\height}{%
            \includegraphics[width=0.65\textwidth]
            {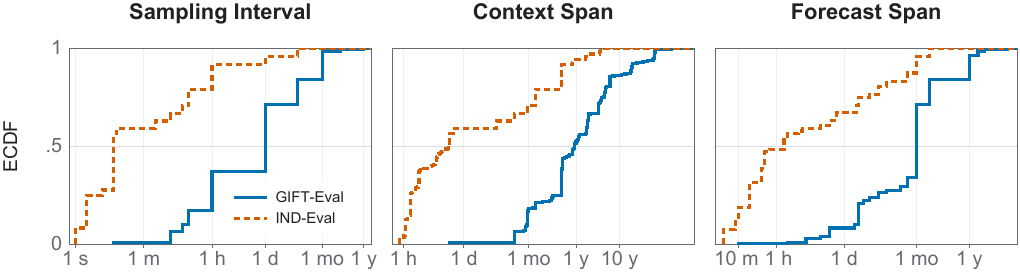}%
        }%
        \hspace{2mm}%
        \raisebox{-0.5\height}{%
            \includegraphics[
                width=\dimexpr0.35\textwidth-2mm\relax
            ]{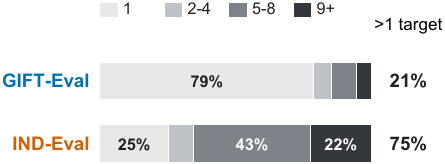}%
        }%
    }

    \caption{Structural coverage of IND-Eval and GIFT-Eval.
    \textbf{Left:} Source-balanced empirical cumulative distribution functions
    of sampling interval, context span, and forecast span.
    \textbf{Right:} Source-balanced distribution of the number of target variates per task.}
    \label{fig:benchmark-structure}
    \captionof{table}{IND-Eval results for Chronos-2, TiRex-2, and
      TimesFM-3 over
      187 tasks. Bold marks the best model per source and metric.
      \textbf{Aggregate} weights all nine sources equally.}
    \label{tab:indeval-results}
    \begingroup
    \scriptsize
    \renewcommand{\arraystretch}{0.78}
    \setlength{\tabcolsep}{3pt}
    \resizebox{\linewidth}{!}{%
      \begin{tabular}{llrrrrrrrrrr}
        \toprule
        Model & Metric & \textbf{Aggregate} & 3W & NEATS & PRONTO & Mesa del Sol & Tablet cmp. & MetroPT-3 & ETT & Electricity & Solar \\
        \midrule
        \multirow{2}{*}{Chronos-2}
          & nMASE & .755 & .847 & .915 & .754 & .864 & .744 & .409 & .853 & \textbf{.680} & .891 \\
          & nCRPS & .539 & .749 & .516 & .569 & .948 & .324 & .322 & .674 & \textbf{.545} & \textbf{.478} \\
        \cmidrule(lr){1-12}
        \multirow{2}{*}{TiRex-2}
          & nMASE & .779 & .837 & \textbf{.899} & .957 & .901 & .736 & .394 & .856 & .723 & .911 \\
          & nCRPS & .537 & .786 & .551 & .719 & .542 & .321 & .322 & .692 & .590 & .523 \\
        \cmidrule(lr){1-12}
        \multirow{2}{*}{TimesFM-3}
          & nMASE & \textbf{.713} & \textbf{.756} & .924 & \textbf{.617} & \textbf{.841} & \textbf{.729} & \textbf{.354} & \textbf{.836} & .699 & \textbf{.869} \\
          & nCRPS & \textbf{.473} & \textbf{.690} & \textbf{.494} & \textbf{.473} & \textbf{.481} & \textbf{.318} & \textbf{.277} & \textbf{.649} & .558 & .482 \\
        \bottomrule
      \end{tabular}%
    }
    \endgroup
\end{figure*}

\paragraph{Benchmark coverage.}\label{subsec:coverage}
Figure~\ref{fig:benchmark-structure} compares the structural coverage of
IND-Eval and GIFT-Eval. IND-Eval consists of datasets with finer sampling rates,
resulting in shorter physical context and forecast durations for a fixed
time-step budget: core horizons span five
minutes to two days, whereas most GIFT-Eval forecast spans exceed one day.
Multivariate forecasting is also substantially more prevalent: the source-level
share of multivariate tasks averages 75\% in IND-Eval, versus 21\% in
GIFT-Eval. These differences characterize structural coverage rather than task
difficulty, and the benchmarks overlap substantially in generic signal
statistics (Appendix~\ref{app:gift-comparison}).

\paragraph{Results.}\label{subsec:results}
We evaluate released zero-shot Chronos-2, TiRex-2, and TimesFM-3
checkpoints under a common IND-Eval protocol that supplies at most 4{,}096
context steps and uses source-balanced nMASE and nCRPS following GIFT-Eval
conventions. Scores divide model error by
seasonal-naive error, so values below one outperform seasonal naive. Geometric
means aggregate tasks within each source and then weight sources equally.
Metrics, baselines, and model-specific inference settings are detailed in
Appendix~\ref{app:metrics}.\\
TimesFM-3 has the lowest source-balanced nMASE (0.713) and nCRPS (0.473),
leading seven of nine sources under each metric.
TiRex-2 leads NEATS nMASE; Chronos-2 leads Electricity nMASE and Electricity
and Solar nCRPS. Some differences are small---Solar nCRPS differs by only
0.004---so we do not treat that ordering as evidence of a meaningful
difference.\\
Mean pairwise absolute differences across sources are 0.18 for nMASE and 0.19
for nCRPS, while the corresponding model differences within sources are 0.05
and 0.08. Descriptively, source-to-source variation exceeds the typical model
gap. Together with the source-specific ranking reversals, this reinforces that
model selection should be aligned with the intended use case rather than a
benchmark-wide ranking. Our provenance audit found no documented pretraining
overlap for the six industrial-core sources, though incomplete corpus
disclosure prevents ruling it out; the core-only aggregate preserves the model
ordering under both metrics (Appendix~\ref{app:contamination}).

\section{Statistic-guided simulation case study}
\label{sec:simulation}
\suppressfloats[t]

\paragraph{Target setting and hypothesis.}
We test an exploratory workflow: diagnose a simulator's target-domain coverage
gaps with simple statistics, add a mechanism intended to reduce them, and test
whether pre-training on adapted simulation improves target forecasts. We study
3W, a multivariate offshore well-monitoring benchmark with labeled
flow-assurance events. The selected severe-slugging (ev3) and flow-instability
(ev4) windows exhibit quasi-periodic dynamics with strong temporal persistence
and low spectral entropy. We compare
SarSim0, SarSim0-oscillator---the coupled-oscillator recipe driven by
SarSim0-generated forcing---and an untuned equal-weight (50/50) mixture used to
probe the specialization--generality trade-off. Selection of
ev3/ev4 and the oscillator recipe was informed by target-context statistics and
preliminary event-level forecasts; no independent design holdout was used
(Appendix~\ref{app:sim-method}). All three use
the same channel-independent PatchTST-based
architecture~\citep{nie2023patchtst} and share the same training budget (Appendix~\ref{app:sim-arch}).
Each model forecasts channels independently; the case study
tests channel-wise temporal alignment, not the cross-channel structure that
IND-Eval also emphasizes.

\paragraph{Diagnosing the gap and choosing a mechanism.}
Simple per-window statistics diagnose SarSim0's largest gaps with 3W:
roughness, spectral entropy, and temporal persistence
(Appendix~\ref{app:sim-stats}). These inexpensive, interpretable summaries
motivate adding smooth, persistent, quasi-periodic dynamics.
Coupled harmonic oscillators provide one simple mechanism for these dynamics:
forced, damped linear elements combine normal modes into smooth, multiscale
responses with resonance and phase lag~\citep{georgi1993physics}. We drive
random planar spring networks with SarSim0-generated forcing and treat projected
displacements at a few readout nodes as synthetic sensor traces
(Appendix~\ref{app:oscillator}, Figure~\ref{fig:oscillator-network}).
The planar geometry is a convenient abstraction for spatially arranged
equipment, not a calibrated model of a particular asset.

\begin{figure}[t]
  \centering
  \begin{minipage}[t]{0.49\linewidth}
    \vspace{0pt}
    \centering
    \includegraphics[width=\linewidth]{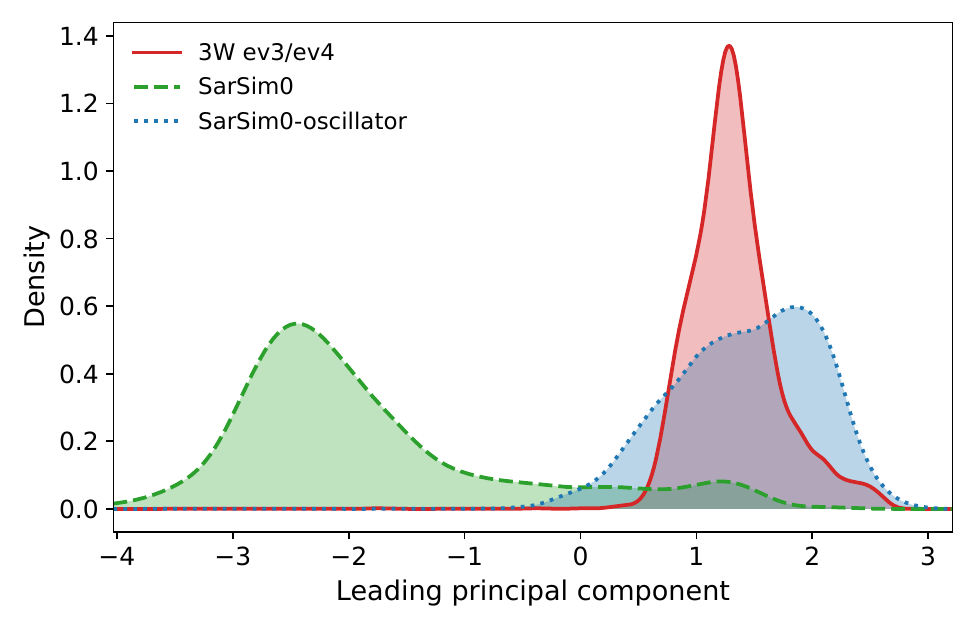}
    \caption{Density along the leading principal component (34\% explained
      variance) of ten standardized statistics for 3W ev3/ev4 and simulated
      windows. SarSim0-oscillator shows substantially greater overlap with 3W
      than SarSim0 does.}
    \label{fig:stat-pca}
  \end{minipage}\hfill
  \begin{minipage}[t]{0.49\linewidth}
    \vspace{0pt}
    \captionof{table}{Downstream forecasting on 3W and the native-univariate
      GIFT-Eval subset. In (b), \# nMASE$<1$ counts tasks outperforming
      seasonal naive.}
    \label{tab:sim-forecast-summary}
    \centering
    \scriptsize
    \textbf{(a) 3W event forecasting}\par
    \vspace{0.5mm}
    \setlength{\tabcolsep}{3pt}
    \begin{tabular}{llrr}
      \toprule
      Event & Recipe & nMASE$\downarrow$ & nCRPS$\downarrow$ \\
      \midrule
      ev3
        & SarSim0 & 0.703 & 0.576 \\
        & SarSim0-oscillator & 0.668 & 0.576 \\
        & 50/50 mixture & \textbf{0.639} & \textbf{0.539} \\
      \specialrule{0.25pt}{1pt}{1pt}
      ev4
        & SarSim0 & 0.714 & 0.471 \\
        & SarSim0-oscillator & 0.698 & 0.459 \\
        & 50/50 mixture & \textbf{0.677} & \textbf{0.437} \\
      \bottomrule
    \end{tabular}
    \par\vspace{1.5mm}
    \textbf{(b) GIFT-Eval native-univariate subset (54/97 tasks)}\par
    \vspace{0.5mm}
    \setlength{\tabcolsep}{3pt}
    \begin{tabular}{lrr}
      \toprule
      Recipe & nMASE$\downarrow$ & \# nMASE$<1$ \\
      \midrule
      SarSim0 & \textbf{0.794} & \textbf{43/54} \\
      SarSim0-oscillator & 1.137 & 21/54 \\
      50/50 mixture & 0.826 & 42/54 \\
      \bottomrule
    \end{tabular}
  \end{minipage}
\end{figure}

\paragraph{Statistic- and shape-space overlap.}
Figure~\ref{fig:stat-pca} compares windows from 3W ev3/ev4 with windows
generated by SarSim0 and SarSim0-oscillator along the leading principal
component of ten jointly standardized per-window statistics
(Appendix~\ref{app:sim-method}).
Along this 34\%-variance component, SarSim0-oscillator overlaps substantially
more with 3W than SarSim0. The match remains partial: autocorrelation and
spectral entropy move closer, while roughness and trend-strength gaps remain
(Appendix~\ref{app:sim-stats}). Because agreement on selected statistics could
miss other structure, we also compare windows using log-band power spectra,
which summarize spectral structure while discarding Fourier phase. A simulated
window is the nearest neighbor of 17.0\% of 3W windows for
SarSim0-oscillator versus 2.5\% for SarSim0; resampled shapes and
autocorrelations give the same ordering
(Appendix~\ref{app:sim-raw-tsne}, Table~\ref{tab:window-shape-check}).

\paragraph{Downstream forecasting and generality.}
To test whether the improved coverage transfers to forecasting, we use one
training run per recipe and evaluate the resulting models on the IND-Eval 3W
tasks for ev3 and ev4. For cross-domain comparison, we use the 54 of 97
GIFT-Eval tasks with natively univariate targets. This matches the
channel-independent evaluation protocol and avoids presenting decomposed
multivariate tasks as evidence of multivariate performance
(Table~\ref{tab:sim-forecast-summary}).

Relative to SarSim0, SarSim0-oscillator has lower nMASE on both events, nCRPS
tied on ev3, and lower nCRPS on ev4. In this
exploratory comparison, greater measured overlap accompanies lower forecasting
error on these events, consistent with statistic-guided simulator design.
SarSim0 remains strongest
on the native-univariate GIFT-Eval subset (nMASE 0.794), showing the cost of
specialization. The 50/50 mixture has the lowest reported error on all four 3W
event--metric pairs and recovers most of the performance lost by
SarSim0-oscillator on the subset (nMASE 0.826 versus 1.137).

\section{Conclusion}
\label{sec:conclusion}

We introduced IND-Eval, a public benchmark of source-specific industrial
forecasting tasks. Across three TSFMs, TimesFM-3 led aggregate performance, but
source-specific ranking reversals showed the need for use-case-aligned model
selection. This finding motivated a case study testing whether target
statistics can guide simulator design when representative training data are
scarce.

We focused on severe slugging and flow instability in 3W, whose windows were
poorly matched by SarSim0 on the selected statistics and representations. The
oscillator-augmented recipe substantially improved this measured alignment.
Pretraining on the augmented data reduced target forecast error but degraded
performance on the native-univariate GIFT-Eval subset. A 50/50 mixture of
generic and augmented data achieved the lowest target errors while recovering
most of the lost GIFT-Eval performance.

These exploratory results link domain-representative evaluation to model
selection and simulator design, exposing a specialization--generality
trade-off.

\begin{ack}
  We thank Niek Beckers for proofreading the paper and helping to finalize
  the manuscript. No third-party funding supported this work. The authors
  declare no competing interests beyond the affiliation above.
\end{ack}

\bibliographystyle{plainnat}
\bibliography{references}

\appendix

\section{Dataset statistics}
\label{app:stats}

\subsection{IND-Eval sources and task construction}
\label{app:ind-composition}

We preprocess all nine datasets into the same context--target schema. A task
combines one source-specific series group with one forecast horizon. Source
instances and forecast windows remain separately identified within each task.
The standard context cap supplied to the model is 4{,}096 samples. Most
industrial sources use 512- and 1{,}024-sample horizons; the batch-oriented and
GIFT-Eval-derived sources retain horizons that match their original evaluation
settings. Every forecast window remains within one source instance. Cadence or
data gaps in NEATS, Mesa del Sol, and MetroPT-3 and batch or active-run breaks
in tablet compression define instance boundaries rather than being spanned.
This guarantee does not prohibit within-instance operating transitions:
PRONTO places the forecast origin at an annotated event, 3W event~1 stitches a
transient interval to a steady interval before windowing, and whole-instance 3W
tasks may retain class transitions.

For the ev3/ev4 case study, preprocessing does not filter on source instance
type: event 3 inputs include operational and source-simulated instances, while
event 4 inputs are operational; neither event includes hand-drawn instances.

Within this hierarchy, a \emph{source instance} is one independent run,
gap-bounded segment, batch, meter, plant, or other source-native recording. A
\emph{forecast window} is one context--target split at one forecast origin. All
target variates from an instance share that split and are forecast jointly. A
\emph{scored series--window pair} is one target variate within a forecast
window; metrics pool errors over these pairs and their horizon steps.

\begin{table*}[thbp]
  \caption{Current IND-Eval source and task construction. Each task combines
    one source-specific series group and forecast horizon.}
  \label{tab:indeval-source-construction}
  \centering
  \small
  \begin{tabular}{@{}
    >{\raggedright\arraybackslash}p{0.11\textwidth}
    >{\raggedright\arraybackslash}p{0.36\textwidth}
    >{\raggedright\arraybackslash}p{0.385\textwidth}r@{}}
    \toprule
    Source & Recorded system & Task construction & Tasks \\
    \midrule
    NEATS & Offshore drilling streams with up to 12 canonical drilling
      variables at 1, 5, 10, or 12-second cadence & 21 well--event groups,
      each evaluated at 512- and 1{,}024-sample horizons & 42 \\
    3W & Multivariate oil-well instances organized into ten operating-event
      groups & One task per event and 512- or 1{,}024-sample horizon & 20 \\
    PRONTO & Six annotated multiphase-flow transients with 12 targets and
      three known-future control variables & One task per annotated event and
      512- or 1{,}024-sample horizon & 12 \\
    Mesa del Sol & Fifteen months of microgrid power, voltage, frequency, and
      temperature measurements & Gap-bounded segments at 10-second, 2-minute,
      and 5-minute resolutions; 512-sample horizon & 3 \\
    Tablet compression & Commercial batches with five direct-compression
      process targets & 25 product-code groups at 30-, 60-, and 180-sample
      horizons; one shared origin per batch & 75 \\
    MetroPT-3 & Five channels from a metro train air-production unit & 190
      leak-free, gap-bounded instances at 10-second cadence, evaluated at 30-,
      60-, and 180-sample horizons & 3 \\
    ETT & Two seven-channel electricity-transformer series & GIFT-Eval
      variants at 15-minute, hourly, daily, and weekly cadences & 16 \\
    Electricity & 370 client electricity meters, treated as independent
      univariate series & GIFT-Eval variants at 15-minute, hourly, daily, and
      weekly cadences & 8 \\
    Solar & Simulated power from 137 photovoltaic plants, treated as
      independent univariate series & GIFT-Eval variants at 10-minute, hourly,
      daily, and weekly cadences & 8 \\
    \midrule
    \multicolumn{3}{r}{Total} & 187 \\
    \bottomrule
  \end{tabular}
\end{table*}

\begin{table*}[thbp]
  \caption{Source-specific preprocessing, forecast-origin selection, and
    exclusions. Rules that inspect the target interval are stated explicitly;
    remaining implementation details and exact source versions are included
    with the released preprocessing scripts.}
  \label{tab:indeval-preprocessing}
  \centering
  \scriptsize
  \renewcommand{\arraystretch}{1.05}
  \begin{tabular}{@{}
    >{\raggedright\arraybackslash}p{0.105\textwidth}
    >{\raggedright\arraybackslash}p{0.43\textwidth}
    >{\raggedright\arraybackslash}p{0.405\textwidth}@{}}
    \toprule
    Source & Cleaning and resampling & Forecast-origin and exclusion rule \\
    \midrule
    NEATS & Canonicalize channel aliases, units, and timestamps; retain up to
      12 drilling targets; split at every cadence discontinuity; remove a
      channel if it is more than 80\% missing, has at most three distinct
      values, or is flat on more than 75\% of valid transitions. & Take up to
      five forward-anchored windows per source instance, 2{,}048 samples apart.
      An instance contributes only to horizons for which a full context and
      target fit. \\
    3W & Apply event-specific extraction, including a stitched transient and
      steady interval for event 1; mean-downsample to 2-second cadence except
      event 4, which remains at 1 second; interpolate sensor gaps; remove
      sparse, frozen, spike-dominated, and low-range channels. & Take up to two
      forward-anchored windows per instance. Except for event 9, drop a window
      only if every retained channel's final 1{,}000 context points plus target
      span less than 5\% of its earlier context range. This rule inspects the
      target interval. \\
    PRONTO & Rename and align 15 process channels, then average the native 1-Hz
      data into 2-second bins. Twelve response channels are targets; Air In,
      Water In, and Air Valve are known-future covariates. & Fix origins at the
      six annotated events, use up to 4{,}096 pre-event samples, and skip an
      event--horizon pair when its pre-event history or post-event target is
      shorter than the horizon. \\
    Mesa del Sol & Mask telemetry sentinels, known invalid codes, and short
      negative spikes; aggregate most channels by mean and one setpoint by its
      last value at 10-second, 2-minute, and 5-minute cadence. Mean bins require
      80\% native-sample coverage. Remove two near-constant setpoint channels
      and split at fully missing bins. & Take every non-overlapping 512-step
      window that fits within each gap-bounded segment. \\
    Tablet compression & Preserve acquisition order; split when the batch or
      active-production state changes, time moves backwards, or a gap occurs;
      mask physically invalid \texttt{SREL}>15. Five process channels are
      targets. & Choose the earliest origin with at least 512 active context
      rows and a finite 180-step future for every target. Reuse that origin for
      the 30-, 60-, and 180-step horizons. This rule inspects the longest target
      interval. \\
    MetroPT-3 & Sort by timestamp, remove four reported leak intervals, and
      split at the resulting row discontinuities or timestamp gaps over 15
      seconds. No value resampling, interpolation, imputation, or scaling is
      applied to the output. & Use the latest origin with a shared 180-step
      future and 512--4{,}096 context samples. Remove dead-tail instances and
      require mean normalized context variability of at least 0.60 across four
      activity channels, with at least two channels at 0.10 or above. The
      dead-tail rule inspects the target interval. \\
    ETT & Keep all seven targets at 15-minute cadence; construct hourly, daily,
      and weekly variants by consecutive sums and drop incomplete trailing
      bins, matching GIFT-Eval. & Use GIFT-Eval's tail-anchored origins, horizon
      counts, and terms. For horizons above the model's native limit, retain
      extra stored history so inference-time block downsampling still supplies
      4{,}096 model steps. \\
    Electricity & Keep GIFT-Eval's four cadence variants and treat the 370
      meters as independent univariate instances. Pre-installation values
      remain missing rather than being imputed. & Use GIFT-Eval's tail-anchored
      origins and terms. Drop a window when more than 60\% of its context and
      target values are missing. \\
    Solar & Keep GIFT-Eval's four cadence variants and treat the 137 plants as
      independent univariate instances. & Use GIFT-Eval's tail-anchored origins
      and terms, with the same stored-context adjustment as ETT for horizons
      above the model's native limit. \\
    \bottomrule
  \end{tabular}
\end{table*}

For NEATS, 3W, Mesa del Sol, ETT, Electricity, and Solar, we discard a
candidate window when more than 60\% of its pooled context and target cells are
missing; otherwise, missing values remain for the forecasting model to handle.
PRONTO, tablet compression, and MetroPT-3 instead use the source-specific
origin and exclusion rules in Table~\ref{tab:indeval-preprocessing}. Apart from
the 3W interpolation listed there, preprocessing does not impute missing sensor
values. Inputs are not scaled before inference; MetroPT-3 uses global standard
deviations only to decide whether a context is sufficiently active.

\begin{table}[thbp]
  \caption{Cadence, forecast horizon, and seasonal-naive period per source.
    Physical horizons are horizon samples times cadence; seasonal periods are
    measured in samples and follow the cadence order shown in each row.}
  \label{tab:indeval-cadence}
  \centering
  \small
  \begin{tabular}{@{}l l l l l@{}}
    \toprule
    Source & Cadence & Horizon (samples) & Horizon (physical) & Seasonal period $m$ \\
    \midrule
    NEATS & 1, 5, 10, 12\,s & 512, 1{,}024 & 8.5\,min--3.4\,h & 1 \\
    3W & 2\,s (event 4: 1\,s) & 512, 1{,}024 & 8.5--34\,min & 1 \\
    PRONTO & 2\,s & 512, 1{,}024 & 17--34\,min & 1 \\
    Mesa del Sol & 10\,s, 2\,min, 5\,min & 512 & 85\,min, 17\,h, 43\,h & 1, 720, 288 \\
    Tablet compression & 10\,s (nominal) & 30, 60, 180 & 5--30\,min & 1 \\
    MetroPT-3 & 10\,s & 30, 60, 180 & 5--30\,min & 1 \\
    ETT & 15\,min, 1\,h, 1\,d, 1\,w & 48/480/720; 30; 8 & 12\,h--30\,d; 8\,w & 96, 24, 1, 1 \\
    Electricity & 15\,min, 1\,h, 1\,d, 1\,w & 48/480/720; 30; 8 & 12\,h--30\,d; 8\,w & 96, 24, 1, 1 \\
    Solar & 10\,min, 1\,h, 1\,d, 1\,w & 48/480/720; 30; 8 & 8\,h--30\,d; 8\,w & 144, 24, 1, 1 \\
    \bottomrule
  \end{tabular}
\end{table}

\paragraph{Cadence, horizon, and seasonal period.}
Table~\ref{tab:indeval-cadence} lists, per source, the sampling cadence, the
forecast horizons in samples and physical time, and the seasonal period $m$ in
samples used by seasonal naive and MASE scaling. Core horizons range from five
minutes (MetroPT-3, tablet compression) to roughly two days (Mesa del Sol at
5-minute resolution); the shared sources keep GIFT-Eval's horizons of 48, 480,
and 720 samples at their two finest cadences and 30 and 8 samples at daily and
weekly cadence. For the industrial sources, we use $m>1$ only when a physical
or calendar cycle applies across a task, fits within the available context,
and has autocorrelation or spectral support. Otherwise $m=1$, so seasonal naive
repeats the most recent observation. Mesa del Sol retains a daily period at
2- and 5-minute resolution, where the context contains several days of data.
At 10-second resolution, one day spans 8{,}640 samples and exceeds the
4{,}096-sample context, so we use $m=1$. We also use $m=1$ for 3W, NEATS,
PRONTO, tablet compression, and MetroPT-3 because their runs, batches, or
variable operating cycles do not support one fixed period. The GIFT-Eval-derived
ETT, Electricity, and Solar tasks retain their original periods.

\paragraph{Evaluation composition.}
The following counts are scored series--window pairs, not joint
multivariate windows. Thus one multivariate window contributes one pair per
target variate. For example, Mesa del Sol's 7{,}701 joint windows across 15
targets yield 115{,}515 scored pairs. Across the current nine-source benchmark,
Mesa del Sol contributes 115{,}515 of 200{,}757 pairs (58\%) and Electricity
contributes 37{,}117 (18\%). Tablet compression, 3W, NEATS, and Solar contribute
13{,}785, 11{,}429, 9{,}934, and 8{,}905 pairs, respectively (4--7\% each);
MetroPT-3, ETT, and PRONTO together contribute about 2.0\% (2{,}850, 1{,}078,
and 144 pairs). Table~\ref{tab:indeval-source-construction} reports task counts
instead. The statistics in Section~\ref{app:gift-comparison} weight every
source dataset equally and are therefore unaffected by this imbalance. The
aggregate row of Table~\ref{tab:indeval-results} is similarly insensitive to
it: both metrics weight the nine sources equally, so no source's contribution
to the aggregate scales with its number of scored pairs
(Section~\ref{app:metrics}).

\subsection{Comparison with GIFT-Eval}
\label{app:gift-comparison}

\begin{figure*}[thbp]
\centering
\includegraphics[width=\textwidth]{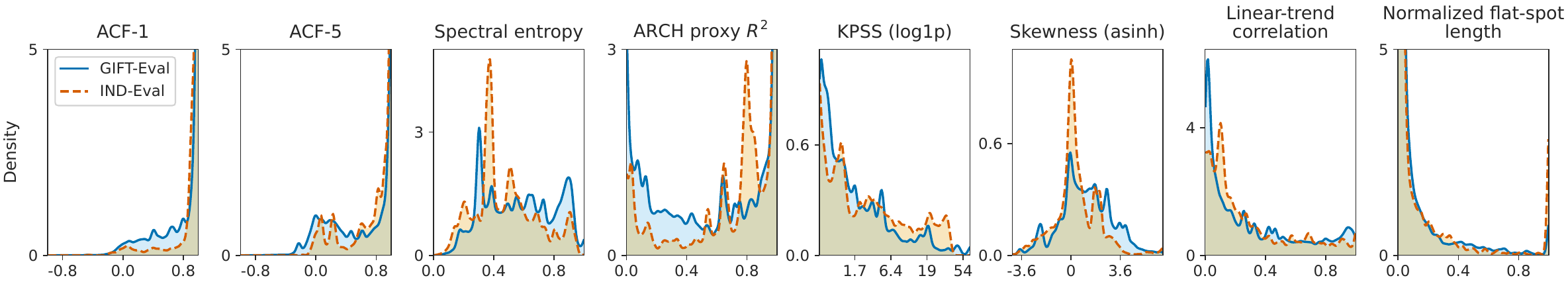}
\caption{Source-balanced densities of the eight benchmark-comparison
statistics: ACF-1, ACF-5, spectral entropy, ARCH proxy $R^2$, KPSS, skewness,
linear-trend correlation, and normalized flat-spot length. Within each
benchmark, every source dataset contributes equal total weight. KPSS and
skewness use $\log(1+x)$ and $\mathrm{asinh}(x)$ scales, respectively, with
tick labels shown in the original units.}
\label{fig:gift-ind-statistics}
\end{figure*}

We compare eight source-balanced marginal statistics on evaluation windows
from GIFT-Eval and IND-Eval. Figure~\ref{fig:gift-ind-statistics} shows their
distributions, and Table~\ref{tab:gift-ind-medians} reports the corresponding
medians. This benchmark-level comparison is separate from the ten-statistic
representation used for the simulation case study in
Appendix~\ref{app:sim-method}.

\paragraph{Statistic definitions.}
ACF-$k$ is the Pearson correlation between paired $x_t$ and $x_{t-k}$ values.
Spectral entropy is $-\sum_{j=1}^{m}p_j\log(p_j)/\log(m)$ for normalized
non-DC periodogram powers $p_j$~\citep{goerg2013forecastable}. ARCH proxy $R^2$
is the coefficient of determination from regressing squared demeaned values on
an intercept and 12 own lags. It is the auxiliary-regression $R^2$ underlying
the ARCH-LM procedure, not its $nR^2$ test
statistic~\citep{engle1982autoregressive}. KPSS is the level-stationarity
statistic with a constant and
$\lfloor4(n/100)^{1/4}\rfloor$ lags~\citep{kwiatkowski1992testing}; larger
values provide stronger evidence against level stationarity. Skewness is the
bias-corrected Fisher--Pearson coefficient. Linear-trend correlation is
$|\mathrm{corr}(t,x_t)|$. Following the binned flat-spot construction of
\citet{hyndman2015large}, normalized flat-spot length is the longest
consecutive run in one of ten equal-width value bins divided by $n$.

The distributions overlap substantially, but IND-Eval tends toward higher
autocorrelation and squared-value lag predictability, as measured by ARCH proxy
$R^2$, and toward lower spectral entropy and skewness. Its KPSS distribution
also extends toward higher values, while linear-trend correlation and
normalized flat-spot length are more similar. This overlap motivates the
structural comparison in Section~\ref{sec:indeval}: marginal statistics alone
do not capture differences in sampling resolution, context and forecast
duration, target dimensionality, or application domain.

\begin{table}[thbp]
    \centering
    \small
    \caption{Source-balanced medians for the eight statistics in
    Figure~\ref{fig:gift-ind-statistics}. Each source dataset contributes equal
    total weight within its benchmark; values are in the original units.}
    \label{tab:gift-ind-medians}
    \begin{tabular}{lcc}
        \toprule
        Statistic & GIFT-Eval & IND-Eval \\
        \midrule
        ACF-1              & 0.895 & 0.960 \\
        ACF-5              & 0.555 & 0.823 \\
        Spectral entropy   & 0.559 & 0.428 \\
        ARCH proxy $R^2$              & 0.667 & 0.845 \\
        KPSS                          & 1.061 & 2.164 \\
        Skewness                      & 0.581 & 0.091 \\
        Linear-trend correlation      & 0.187 & 0.171 \\
        Normalized flat-spot length   & 0.034 & 0.028 \\
        \bottomrule
    \end{tabular}
\end{table}

\subsubsection{Structural coverage figure}
\label{app:structural-coverage}

Figure~\ref{fig:benchmark-structure} compares the structural coverage of
IND-Eval and GIFT-Eval. For each task configuration, we denote the sampling
interval by $\Delta$, the effective context length by $C$, and the forecast
horizon by $H$. We compute the context and forecast spans as $C\cdot\Delta$
and $H\cdot\Delta$, respectively. IND-Eval provides pre-windowed tasks.
For GIFT-Eval, we first construct evaluation instances using its official
rolling-window procedure and then cap the context supplied to the model at
4,096 points, matching the source-balanced median context length in IND-Eval.

To prevent sources with more series or task variants from dominating the
comparison, we assign each source dataset equal total weight and divide it
equally among its distinct $(\Delta,C,H)$ configurations. The variate
distribution uses analogous source balancing and reports the number of
target variates forecast jointly.

\section{Metrics and inference settings}
\label{app:metrics}

\subsection{Normalized scores}
\label{app:normalized-scores}

For each task, nMASE and nCRPS divide the model's MASE and CRPS, pooled over
all scored series--window pairs and horizon steps of the task, by the
corresponding score of the seasonal-naive forecast with the seasonal period of
Table~\ref{tab:indeval-cadence}. At the task level, a ratio of one indicates
the same error as seasonal naive under the corresponding metric; values below
one indicate lower error, and values above one indicate higher error. Per
source, nMASE and nCRPS are geometric means over the source's tasks. The
aggregate row of Table~\ref{tab:indeval-results} then weights sources equally,
again using the geometric mean.

\subsection{Naive baselines}
\label{app:baselines}

We implement both persistence baselines with Nixtla's Python
\texttt{StatsForecast}
\texttt{SeasonalNaive}~\citep{garza2022statsforecast}. Setting its period to
$m=1$ gives the naive forecast, which repeats the latest observation; using the
task-specific $m$ from Table~\ref{tab:indeval-cadence} repeats the final
observed season. Before fitting, we replace missing context values by
last-value imputation. The median is the resulting persistence forecast. Other
requested quantiles are the lower or upper endpoints of symmetric,
residual-based prediction intervals with nominal coverage $2|q-0.5|$ for
quantile $q$. They therefore come from the baseline's fitted residuals, not a
separate probabilistic forecasting model.

Table~\ref{tab:indeval-baselines} reports the naive (last-value) and
seasonal-naive baselines under the metrics of Table~\ref{tab:indeval-results}.
Seasonal naive has nMASE and nCRPS of one by construction. The two baselines
coincide whenever $m=1$: for every 3W, NEATS, PRONTO, tablet-compression, and
MetroPT-3 task; for 10-second Mesa del Sol tasks; and for the daily and weekly
ETT, Electricity, and Solar tasks. The aggregate row is recomputed from the
per-source values with the rule of Appendix~\ref{app:metrics}.

\begin{table}[thbp]
  \caption{Naive and seasonal-naive baselines on IND-Eval. Sources are ordered
    as in Table~\ref{tab:indeval-results}; \textbf{Aggregate} weights all nine
    sources equally.}
  \label{tab:indeval-baselines}
  \centering
  \begingroup
  \small
  \renewcommand{\arraystretch}{0.9}
  \setlength{\tabcolsep}{3pt}
  \begin{tabular*}{\linewidth}{@{\extracolsep{\fill}}l rr rr@{}}
    \toprule
    & \multicolumn{2}{c}{nMASE$\downarrow$}
    & \multicolumn{2}{c}{nCRPS$\downarrow$} \\
    \cmidrule(lr){2-3}\cmidrule(l){4-5}
    Source & Naive & S-naive & Naive & S-naive \\
    \midrule
    \textbf{Aggregate} & 1.217 & 1.000 & 1.294 & 1.000 \\
    \midrule
    3W           & 1.000 & 1.000 & 1.000 & 1.000 \\
    NEATS        & 1.000 & 1.000 & 1.000 & 1.000 \\
    PRONTO       & 1.000 & 1.000 & 1.000 & 1.000 \\
    Mesa del Sol & 1.280 & 1.000 & 1.410 & 1.000 \\
    Tablet cmp.  & 1.000 & 1.000 & 1.000 & 1.000 \\
    MetroPT-3    & 1.000 & 1.000 & 1.000 & 1.000 \\
    \midrule
    ETT          & 1.280 & 1.000 & 1.590 & 1.000 \\
    Electricity  & 2.050 & 1.000 & 2.510 & 1.000 \\
    Solar        & 1.740 & 1.000 & 1.810 & 1.000 \\
    \bottomrule
  \end{tabular*}
  \endgroup
\end{table}

\subsection{Model inference settings}
\label{app:inference}

Table~\ref{tab:inference-settings} summarizes the model-specific protocol used
for Table~\ref{tab:indeval-results}. All three adapters receive at most 4{,}096
context samples, forecast the target variates of a multivariate window jointly,
and treat genuinely univariate instances independently. PRONTO is the only
source with known-future covariates; its three control channels are supplied to
all three models. Missing context values are passed to each model's native
handling without adapter-level imputation. For quantile forecasts, we use the
standard quantile-loss approximation to CRPS:
\[
  \widehat{\mathrm{CRPS}}(F,y)
  = \frac{2}{K}\sum_{k=1}^{K}\rho_{\tau_k}(y-\hat q_{\tau_k}),
  \qquad
  \rho_\tau(u)=u\bigl(\tau-\mathbf{1}\{u<0\}\bigr).
\]
We use $K=9$ levels $\tau_k=0.1,\ldots,0.9$ before the seasonal-naive
normalization of Appendix~\ref{app:normalized-scores}.

No IND-Eval horizon exceeds 1{,}024 samples, so the benchmark's block
downsampling bridge is not used for these results. Checkpoint identifiers were
resolved from Hugging Face without immutable model-weight revision pins; the
identifiers below therefore record the model repositories, not frozen weight
commits. The evaluated TimesFM-3 implementation was the \texttt{timesfm} 3.0.1
package, whose official repository is
\url{https://github.com/google-research/timesfm}; the experiment records do not
preserve the corresponding upstream code commit.

\begin{table*}[thbp]
  \caption{Model-specific inference settings for
    Table~\ref{tab:indeval-results}.}
  \label{tab:inference-settings}
  \centering
  \scriptsize
  \renewcommand{\arraystretch}{1.08}
  \setlength{\tabcolsep}{3pt}
  \begin{tabular}{@{}
    >{\raggedright\arraybackslash}p{0.12\textwidth}
    >{\raggedright\arraybackslash}p{0.27\textwidth}
    >{\raggedright\arraybackslash}p{0.27\textwidth}
    >{\raggedright\arraybackslash}p{0.27\textwidth}@{}}
    \toprule
    Setting & Chronos-2 & TiRex-2 & TimesFM-3 \\
    \midrule
    Checkpoint
      & \texttt{amazon/chronos-2}
      & \texttt{NX-AI/TiRex-2}
      & \texttt{google/timesfm-3.0-pytorch}; \texttt{timesfm} 3.0.1 \\
    Targets
      & Joint within multivariate windows; separate across windows
      & Joint within multivariate windows; separate across windows
      & Joint within multivariate windows; separate across windows \\
    PRONTO controls
      & Split into past and known-future covariate dictionaries
      & Supplied as one context-plus-future covariate tensor
      & Supplied as \texttt{past\_future\_covariates} \\
    Quantiles
      & Requested at 0.1--0.9
      & Checkpoint-native 0.1--0.9
      & Checkpoint-native 0.1--0.9 \\
    Horizon handling
      & Native for all tasks (adapter cap 2{,}048)
      & Chunks above checkpoint \texttt{future\_len}, feeding back the median
        forecast
      & Native for all tasks (adapter cap 2{,}048) \\
    Other settings
      & \texttt{cross\_learning=False}; float32 inference
      & Checkpoint-native forecast routine
      & \texttt{use\_symmetric\_averaging}, \texttt{make\_positive}, and
        \texttt{sort\_quantiles} enabled \\
    \bottomrule
  \end{tabular}
\end{table*}

\section{Pretraining-overlap and provenance audit}
\label{app:contamination}

Table~\ref{tab:indeval-results} reports zero-shot forecasts from Chronos-2
\citep{ansari2025chronos2}, TiRex-2 \citep{podest2026tirex2}, and TimesFM-3
\citep{jain2026timesfm3}. A zero-shot
score cannot by itself establish generalization if the model already saw the
evaluation series during pretraining, so we audit every model--source pair for
overlap.

\paragraph{Method.}
No model documents its pretraining corpus in full, so we cannot match
values against it directly. We audit \emph{provenance} instead. For each
IND-Eval source we record its first public release, whether the source or a
close derivative appears in GIFT-Eval \citep{aksu2024gifteval} or another
widely used forecasting benchmark, and whether the models' reports name the
source as training data. Each pair then gets one of three evidence labels.
\emph{Documented} means official model documentation names the source or a
clearly linked derivative. \emph{Possible} means the same underlying series
could plausibly have entered training through a re-packaged benchmark, a
resampled copy, or the training split of a partly held-out dataset.
\emph{Unresolved} means no documented evidence points either way, which is
absence of evidence rather than evidence of absence. This distinction between
documented exposure and unresolved absence of evidence follows a concurrent
TSFM contamination audit \citep{li2026tsfmaudit}.
IND-Eval does not restrict evaluation windows to source segments that postdate
model data cut-offs. For ETT, Electricity, and Solar, the underlying data end in
2018, 2014, and 2006, respectively.
IND-Eval also releases the raw source series alongside the evaluation schema
(Section~\ref{sec:introduction}), enabling independent value-level overlap
checks whenever a pretraining corpus is available.

\begin{table}[htbp]
  \caption{\textbf{Model--source contamination audit for
    Table~\ref{tab:indeval-results}.} \emph{Rel.}\ is the first verified public
    release year. For NEATS, the records were accessed in 2026 but their first
    public release could not be verified. Prior forecasting benchmarks are
    GIFT-Eval \citep{aksu2024gifteval}, TIME \citep{qiao2026time}, the Monash archive
    \citep{godahewa2021monash}, and the LTSF benchmark
    \citep{zeng2023transformers}. Evidence is \emph{documented} when official
    model documentation names the source or a linked derivative,
    \emph{possible} for plausible indirect overlap, and \emph{unresolved} when
    public documentation does not establish either case. These labels do not
    establish exact sample duplication. No source is evaluated only on a
    segment that postdates a model cut-off.}
  \label{tab:contamination-matrix}
  \centering
  \begingroup
  \small
  \renewcommand{\arraystretch}{1.15}
  \begin{tabular*}{\linewidth}{@{\extracolsep{\fill}}
    >{\raggedright\arraybackslash}p{0.21\linewidth}c
    >{\raggedright\arraybackslash}p{0.19\linewidth}ccc@{}}
    \toprule
     & & & \multicolumn{3}{c}{Pretraining-overlap evidence} \\
    \cmidrule(l){4-6}
    Source & Rel. & Prior forecasting benchmark(s)
      & Chronos-2 & TiRex-2 & TimesFM-3 \\
    \midrule
    \multicolumn{6}{@{}l}{\textit{Industrial core sources and documented
      benchmark overlap}} \\
    3W & 2019 & none & unresolved & unresolved & unresolved \\
    NEATS & unknown & none & unresolved & unresolved & unresolved \\
    PRONTO & 2019 & none & unresolved & unresolved & unresolved \\
    Mesa del Sol & 2023 & none & unresolved & unresolved & unresolved \\
    Tablet compression & 2022 & none & unresolved & unresolved & unresolved \\
    MetroPT-3 & 2022 & TIME (5-min.) & unresolved & unresolved & unresolved \\
    \addlinespace
    \multicolumn{6}{@{}l}{\textit{Shared forecasting sources and
      pretraining-overlap evidence}} \\
    ETT-small & 2021 & GIFT-Eval, LTSF & possible & possible & possible \\
    Electricity & 2015 & GIFT-Eval, Monash, LTSF
      & documented & documented & possible \\
    Solar (Alabama) & 2018 & GIFT-Eval, Monash
      & documented & documented & possible \\
    \bottomrule
  \end{tabular*}
  \endgroup
\end{table}

\paragraph{Model checkpoints.}
We evaluate the released \texttt{amazon/chronos-2} checkpoint
($\sim$120M parameters, released October 2025), the released
\texttt{NX-AI/TiRex-2} checkpoint (38.4M univariate plus 44.1M multivariate
activated parameters, released July 2026), and the released
\texttt{google/timesfm-3.0-pytorch} checkpoint (330M parameters, released
August 2026)~\citep{jain2026timesfm3}.
No report gives a precise corpus-wide data
cut-off date, so we use the checkpoint release dates as a proxy. Chronos-2
states that its corpus was built to exclude the \emph{test} portions of all
GIFT-Eval tasks at every sampling frequency, while keeping partial overlap with
some GIFT-Eval \emph{training} portions. TiRex-2 inherits the TiRex
\citep{auer2025tirex} pretraining mixture, which includes the Chronos training
corpus and the GIFT-Eval-Pretrain collection, and it also publishes
benchmark-specific decontaminated checkpoints such as \texttt{TiRex-2-g}, which
excludes all GIFT-Eval overlap. The results in
Table~\ref{tab:indeval-results} use the base \texttt{TiRex-2} checkpoint.
TimesFM-3 reports pretraining on the GIFT-Eval-Pretrain collection excluding
datasets that overlap fev-bench \citep{shchur2025fev}, web-traffic series, and
synthetic and augmented data~\citep{jain2026timesfm3}; its stated exclusion
therefore targets fev-bench rather than GIFT-Eval.%
\footnote{\url{https://huggingface.co/google/timesfm-3.0-pytorch}}

\paragraph{Findings.}
No model names any of the six industrial-core sources in its disclosed
corpus. Five also do not appear in the forecasting benchmarks reviewed here;
MetroPT-3 is the exception, appearing in TIME after downsampling to five-minute
resolution~\citep{qiao2026time}, whereas IND-Eval retains gap-bounded episodes
at 10-second cadence. We therefore classify all six sources as
\emph{unresolved} for all three models: this is absence of documented evidence,
not evidence that pretraining overlap is absent. By contrast, ETT, Electricity,
and Solar are long-standing GIFT-Eval series. Electricity and Solar are
documented for Chronos-2 and TiRex-2; ETT remains possible because it may occur
in the broader public collections both report. For TimesFM-3 all three are
\emph{possible}: its disclosed corpus does not name them and draws on a
collection documented as non-leaking \citep{aksu2024gifteval}, but that claim
is the collection publisher's own and the augmented component is unspecified.

\paragraph{Effect on aggregates.}
The paper treats source-level results as primary
(Section~\ref{sec:indeval}). As a sensitivity check, recomputing the
source-balanced aggregate from the displayed source scores in
Table~\ref{tab:indeval-results} over only the six industrial-core sources gives
nMASE/nCRPS of 0.674/0.436 for TimesFM-3,
0.732/0.528 for Chronos-2, and 0.757/0.509 for TiRex-2, leaving the model
ordering unchanged on both metrics.
This restriction excludes the three shared sources with documented or possible
pretraining overlap: ETT, Electricity, and Solar. It does not prove that the
retained sources were absent from pretraining because no model documents
its corpus fully.

\section{Simulation case study details}
\label{app:simulation}

\subsection{Harmonic-oscillator augmentation}
\label{app:oscillator}

\begin{figure*}[thbp]
  \centering
  \includegraphics[width=\textwidth]{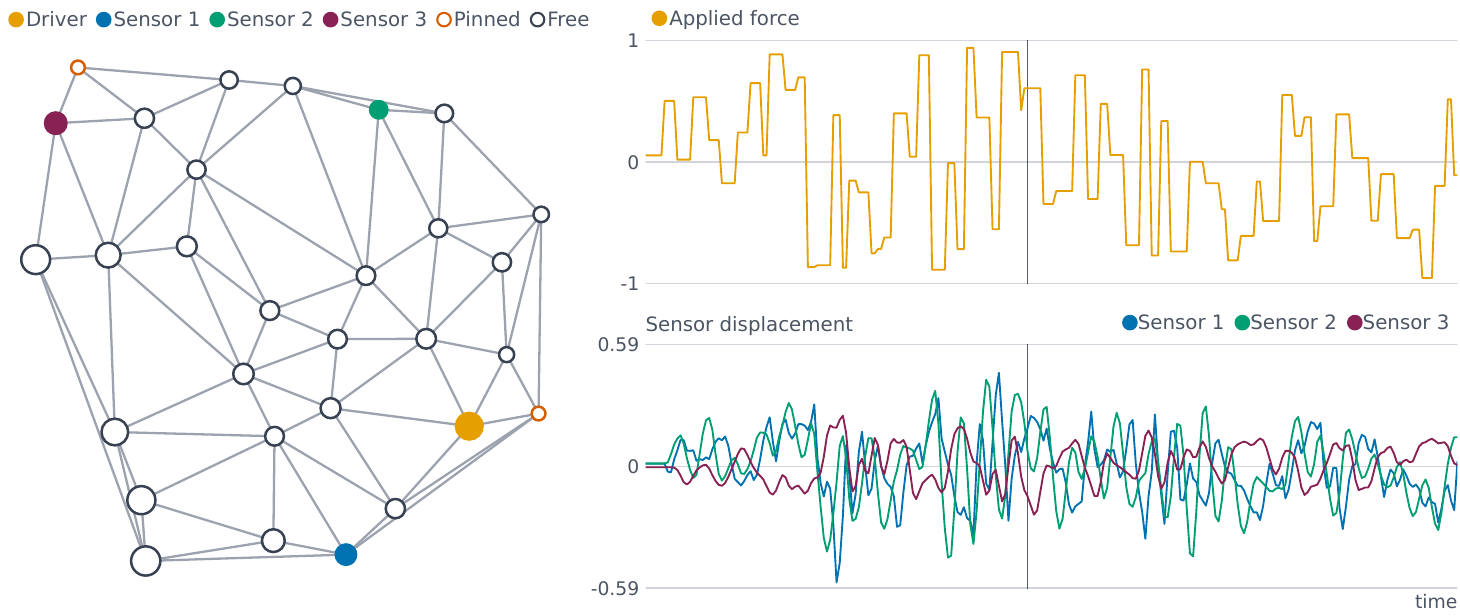}
  \caption{Example harmonic-oscillator augmentation. A directed force drives a
    random planar spring network (left), where node size indicates displacement
    magnitude. Projected displacements at sensor nodes produce rich temporal
    responses (right).}
  \label{fig:oscillator-network}
\end{figure*}

Figure~\ref{fig:oscillator-network} summarizes the harmonic-oscillator
augmentation.

\paragraph{Network and readouts.}
We sample 15--50 well-spaced nodes in a plane and connect their full Delaunay
triangulation with linear axial springs of random positive stiffness. Two
maximally separated convex-hull nodes are pinned in both coordinate directions
to prevent the network as a whole from translating or rotating. We select one
or two unpinned driver nodes, separated in graph distance, and five candidate
readout nodes, preferentially away from the drivers. Each generated sensor set
retains two to five of these readouts. Robustly normalized SarSim0-generated
sequences set the magnitude of forces applied to the drivers along random
directions. Each synthetic sensor trace is a random scalar projection of a
readout node's two-dimensional displacement, rescaled to the robust location
and scale of its inputs.

\paragraph{Dynamics and assumptions.}
For displacement $\mathbf{r}$, stiffness matrix $K$, force-injection matrix
$S$, and driver input $\mathbf{f}$, the unit-mass dynamics are
\begin{equation}
  \ddot{\mathbf{r}} + (\alpha I + \beta K)\dot{\mathbf{r}}
  + K\mathbf{r} = S\mathbf{f}.
\end{equation}
The term $\alpha I+\beta K$ is Rayleigh damping. We sample a target damping
ratio $\zeta\in[0.03,1.5]$ log-uniformly and choose $\alpha,\beta$ so that
$\zeta$ is attained at the slowest and fastest modes. The generator is not
calibrated to a real asset. It assumes planar small-displacement linear springs,
identical unit point masses, proportional viscous damping, dimensionless
parameters, and forces that are piecewise constant between samples.

\paragraph{Exact modal solve.}
After removing pinned degrees of freedom, we diagonalize the reduced symmetric
stiffness matrix. Rayleigh damping preserves this modal basis, reducing the
network to independent damped oscillators. We retain every free mode, choose
the timestep to give 12 samples per period of the fastest mode, and reject
networks whose slowest mode does not complete one cycle in the sequence. Each
force sample is held until the next sample, and cached modal impulse responses
give the exact zero-order-hold response through FFT convolution. The initial
state is the static equilibrium under the first force sample with zero
velocity. Thus there is no modal-truncation or time-stepping approximation
relative to the assumed linear, zero-order-held system. For efficiency, we
reuse the eigensystem and response kernels for up to four independently driven
sensor sets. Each set uses newly selected SarSim0 input rows, force and
projection directions, and two to five of the fixed candidate readout nodes.
Projecting the readouts in modal space avoids constructing the full
node-displacement trajectory.

\paragraph{Univariate and multivariate use.}
The network produces two to five jointly generated sensor traces. For the
univariate encoder in this case study, we discard the force-input rows and
treat each sensor trace as an independent univariate training series. Keeping
the traces together instead yields correlated examples suitable for
multivariate model training.

\subsection{Statistical comparison}
\label{app:sim-method}

\paragraph{Design scope.}
The choice of ev3/ev4 and the oscillator-based recipe was informed by
target-context statistics and preliminary event-level forecast results. The
reported evaluation instances were not reserved from these design decisions,
so this is an exploratory target-informed study rather than a held-out transfer
test. Oscillator parameter ranges are broad simulator defaults rather than
fits to a real asset. We report one equal-weight mixture and one common 50k
checkpoint; no mixture-weight or checkpoint sweep is presented.

For this separate simulator analysis, we compute ten statistics for each 3W
reference or case-study-simulator context: robust skewness, roughness (mean
absolute second difference), trend strength, spectral entropy, ACF level (mean
autocorrelation at lags 1--5), ACF decay (lag 1 minus lag 5), end $z$-score,
isolated spike frequency, recent max $z$-score, and recent level shift.

For 3W, we retain only the 1{,}024-sample-horizon task, yielding 2{,}450
ev3/ev4 windows; the 512-sample-horizon task reuses a subset of the same
contexts. Section~\ref{sec:simulation} nevertheless evaluates both tasks
because their forecast targets differ. Each simulator contributes a fixed
analysis sample of 8{,}192 windows. In the
statistic-space comparison each simulated window is the visible context suffix left by the
training-time left padding of Appendix~\ref{app:sim-arch}, so its length varies
up to 4{,}096 samples, whereas Appendix~\ref{app:sim-raw-tsne} uses the full
simulated context.

Before PCA, we log-transform roughness, then center and scale each statistic
across the pooled sample to zero mean and unit variance by subtracting its mean
and dividing by its standard deviation.

Figure~\ref{fig:stat-pca} visualizes the leading principal component of a
PCA fitted to the pooled standardized vectors.
Appendix~\ref{app:sim-pca-loadings} reports the PCA loadings.

\subsection{Statistic medians}
\label{app:sim-stats}

Table~\ref{tab:sim-all-stats} reports medians for the ten comparison
statistics. SarSim0 differs most from 3W in roughness, spectral entropy, and
temporal persistence. SarSim0-oscillator closely matches the 3W ACF
statistics and moves spectral entropy toward the 3W value, but remains
rougher and more trend-dominated.

\begin{table}[ht]
  \caption{Median values of the ten comparison statistics. Roughness is shown
    before the logarithmic transform used for PCA.}
  \label{tab:sim-all-stats}
  \centering
  \small
  \begin{tabular}{lrrr}
    \toprule
    Statistic & 3W & SarSim0 & SarSim0-oscillator \\
    \midrule
    Robust skewness & $-$0.035 & 0.000 & $-$0.000 \\
    Roughness & 0.0013 & 1.110 & 0.0046 \\
    Trend strength & 0.132 & 0.082 & 0.735 \\
    Spectral entropy & 0.204 & 0.782 & 0.297 \\
    ACF level & 0.999 & 0.386 & 1.000 \\
    ACF decay & 0.002 & 0.682 & 0.001 \\
    End $z$-score & $-$0.046 & $-$0.013 & 0.002 \\
    Isolated spike freq. & 0.000 & 0.000 & 0.000 \\
    Recent max $z$-score & 0.047 & 1.706 & 0.199 \\
    Recent level shift & $-$0.043 & 0.000 & 0.003 \\
    \bottomrule
  \end{tabular}
\end{table}

\subsection{Principal-component loadings}
\label{app:sim-pca-loadings}

Table~\ref{tab:sim-pca-loadings} reports the loadings of the ten standardized
statistics on the leading principal component in
Figure~\ref{fig:stat-pca}, which explains 34\% of the variance.
Because each statistic is standardized before fitting PCA, loading magnitudes
are directly comparable. The component's sign is arbitrary; only relative signs
within it are meaningful. The leading component contrasts ACF level
and trend strength with spectral entropy, ACF decay, and log roughness.

\begin{table}[ht]
  \caption{Loadings of the standardized statistics on the leading principal
  component (34\% explained variance).}
  \label{tab:sim-pca-loadings}
  \centering
  \small
  \begin{tabular}{lr}
    \toprule
    Statistic & Leading component \\
    \midrule
    ACF level & 0.505 \\
    Spectral entropy & $-$0.498 \\
    ACF decay & $-$0.442 \\
    Log roughness & $-$0.384 \\
    Trend strength & 0.369 \\
    Isolated spike freq. & $-$0.093 \\
    Recent max $z$-score & $-$0.090 \\
    Robust skewness & $-$0.020 \\
    End $z$-score & $-$0.012 \\
    Recent level shift & $-$0.011 \\
    \bottomrule
  \end{tabular}
\end{table}

\subsection{Model architecture and training}
\label{app:sim-arch}

\paragraph{Architecture.}
We use a channel-independent PatchTST-based encoder~\citep{nie2023patchtst}:
each series is normalized using its visible-context mean and standard deviation
(RevIN-style~\citep{kim2022revin}), patched with length 64 and stride 32, and
encoded by a 12-layer time-axis transformer
($d_{\mathrm{model}}{=}512$, 8 heads, GELU feed-forward with hidden size
2{,}048).
Rotary position embeddings (RoPE)~\citep{su2024roformer} replace learned
absolute positions; U-Net-style skip connections~\citep{ronneberger2015unet}
and value-residual attention~\citep{zhou2025value} mix early-layer
representations into later layers.
A linear head projects each patch token to 19 quantile forecasts (0.05--0.95);
training minimizes the pinball loss in normalized space.
All 12 layers attend only along time, making this experiment univariate; the
comparison therefore concerns univariate simulator recipes rather than
multivariate cross-row coupling.

\paragraph{Training procedure.}
All three models share identical architecture and optimization; only the
simulation recipe differs. We otherwise follow the SarSim0
defaults~\citep{oreshkin2026zero}. SarSim0 uses the standard observation-noise
mixture, whereas SarSim0-oscillator is driven by passthrough SarSim0 forcing.
The comparison is therefore between complete simulation recipes rather than an
oscillator-only ablation.
The 50/50 mixture samples the two recipes equally. Simulated series have length
$T{=}4{,}608$ (4{,}096 context plus 512 forecast).
We sample a left-padded forecast task with random left padding up to 4{,}088
steps and inject 1\% Bernoulli missingness in the context without loss on
missing cells.
We optimize matrix-valued transformer weights with
NorMuon~\citep{li2026normuon} and remaining parameters with
AdamW~\citep{loshchilov2019adamw}, using learning rate $3{\times}10^{-4}$ and
weight decay 0.01. The effective batch contains 4{,}096 simulated series per
optimizer update, formed from two micro-batches of 2{,}048 by gradient
accumulation.
We compare all recipes after 50k optimizer steps (204.8M simulated series).
Table~\ref{tab:sim-hparams} lists the full hyperparameter settings.
Each of the three recipes was trained on a single H100 or B200 GPU, with
training completing within 24--48 hours. Simulator throughput varied across
recipes and partially bottlenecked training, which accounts for the spread in
wall-clock time; we expect further simulator optimization to narrow this gap.

\begin{table}[t]
  \caption{Model and training hyperparameters for the simulation experiment.
    All recipes use the same settings; only the simulator recipe differs.}
  \label{tab:sim-hparams}
  \centering
  \small
  \begin{tabular}{ll}
    \toprule
    Hyperparameter & Value \\
    \midrule
    Optimizer & NorMuon + AdamW \\
    Learning rate & $3{\times}10^{-4}$ \\
    Weight decay & 0.01 \\
    Training steps reported & 50k \\
    Effective batch size & 4{,}096 (2{,}048 $\times$ 2 accumulation) \\
    Context length & 4{,}096 \\
    Forecast horizon & 512 \\
    Patch length / stride & 64 / 32 \\
    Transformer layers & 12 (time-axis only) \\
    Attention heads & 8 \\
    Transformer width ($d_{\mathrm{model}}$) & 512 \\
    Feed-forward hidden size & 2{,}048 \\
    Activation & GELU \\
    Position encoding & RoPE \\
    Input normalization & per-series mean / std (RevIN-style) \\
    Output quantiles & 19 (0.05--0.95) \\
    U-Net skips & enabled \\
    \bottomrule
  \end{tabular}
\end{table}

\subsection{Window-shape comparison}
\label{app:sim-raw-tsne}

To check whether the statistic-space ordering also appears at the level of whole
windows, we compare 3W reference and case-study-simulator windows under three
per-window representations. We use the same 3W subset and simulator samples as
in Appendix~\ref{app:sim-method}, except that each simulated analysis window
here is the full 4{,}096-step context the simulator generated. The 3W cadence is
2 seconds except for ev4 at 1 second (Table~\ref{tab:indeval-cadence}), and
simulated series carry no physical time unit. We express frequencies in cycles
per sample and lags in samples.

\paragraph{Representations.}
\emph{Shape}: each window is linearly resampled to 256 points and normalized to
zero mean and unit standard deviation. Euclidean distance between such shapes
depends on the timing and phase of features: two windows with the same
oscillation at different phases can be far apart.
Shape distances are dominated by the slow, low-frequency envelope of the windows.
\emph{Log-band power spectrum}: we take the periodogram of
the normalized full-resolution window, average its power within each of 29
log-spaced frequency bands, and take the base-10 logarithm of each band average.
\emph{Autocorrelation}: the sample autocorrelation of the normalized window at
29 log-spaced lags from 1 to 2{,}048 samples. Compared with pointwise shape
distances, the spectral and autocorrelation representations are less sensitive
to phase and emphasize oscillation frequency, roughness, and spectral shape.

\begin{figure*}[thbp]
  \centering
  \includegraphics[width=\textwidth]{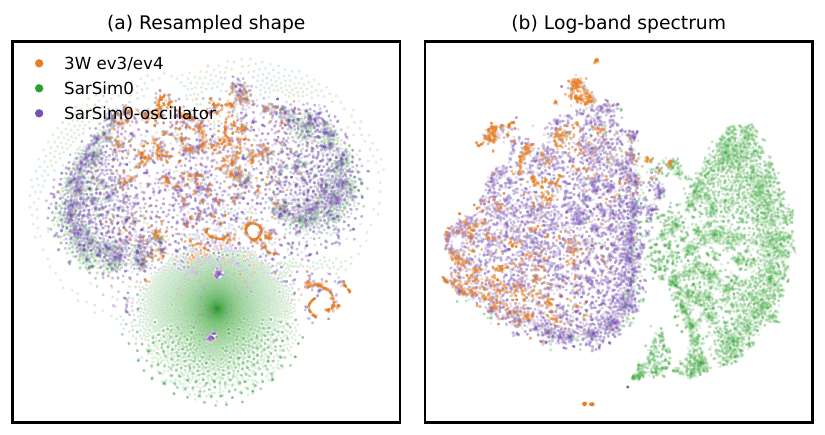}
  \caption{Joint t-SNE projections of windows from 3W ev3/ev4 and both
  simulators in (a) the 256-dimensional resampled-shape space and (b) the
  29-dimensional log-band power spectrum space. The panels are independent
  embeddings of two different representations of the same windows, not
  transforms of one another. Axes and global geometry have no intrinsic
  interpretation.}
  \label{fig:raw-tsne}
\end{figure*}

\paragraph{Metrics.}
We form two pools, each combining the 2{,}450 3W windows with one simulator's
8{,}192 windows, and compute every metric in the feature space itself, with each
coordinate standardized across the pooled windows of all three sources, rather
than on a two-dimensional embedding. The real-to-simulated nearest-neighbor
(NN) rate is the fraction of 3W windows whose nearest other window in the pool
comes from the case-study simulator. Were 3W reference and case-study-simulator
windows exchangeable draws from a single distribution, this rate would be
$8{,}192/(2{,}450+8{,}192-1)=77.0\%$.
The 3W windows are themselves not exchangeable: the 2{,}450 windows come from
427 instances, each contributing one window per retained channel and up to two
time positions, so most 3W windows have close siblings in the 3W reference
sample. We therefore also report a sibling-free NN rate that excludes
candidates from the query window's own instance.

\begin{table}[htbp]
  \centering
  \caption{Window-level alignment of each simulator with 3W ev3/ev4. NN rate is
  the fraction of 3W windows whose nearest neighbor in the pool comes from the
  case-study simulator; the sibling-free variant excludes candidates from the
  query window's own 3W instance.}
  \label{tab:window-shape-check}
  \small
  \begin{tabular}{llrr}
    \toprule
    Representation (dim.) & Simulator & NN rate (\%) & Sibling-free (\%) \\
    \midrule
    Shape (256) & SarSim0            &  6.7 &  7.6 \\
                & SarSim0-oscillator & 19.9 & 27.6 \\
    \addlinespace
    Log-band power (29) & SarSim0            &  2.5 &  3.3 \\
                & SarSim0-oscillator & 17.0 & 24.9 \\
    \addlinespace
    Autocorrelation (29) & SarSim0            &  4.4 &  6.3 \\
                & SarSim0-oscillator & 17.1 & 27.8 \\
    \bottomrule
  \end{tabular}
\end{table}

Table~\ref{tab:window-shape-check} shows the same ordering in all three
representations: SarSim0-oscillator has three to eight times the NN rate of
SarSim0, and excluding sibling windows widens that gap rather than closing it.
The spectral and autocorrelation comparisons support a closer match in spectral
content as well as the low-frequency envelope.

The t-SNE projections in Figure~\ref{fig:raw-tsne} show the same picture. In
shape space, SarSim0 concentrates in a region with few 3W points while many 3W
points intermingle locally with SarSim0-oscillator. In log-band power space the
separation of SarSim0 from 3W is nearly complete, 3W and SarSim0-oscillator
overlap over most of their extent, and a few groups of 3W windows remain
separated from both simulators.

\section{Limitations}
\label{app:limitations}

IND-Eval is a first release covering nine public sources and forecasting only.
We compare three released TSFMs zero-shot against naive baselines, without
fine-tuned or classical baselines; the results therefore characterize these
tasks and models rather than industrial forecasting generally. Task definitions
follow source documentation but have not been validated with asset operators.
Because no model fully discloses its pretraining corpus, our provenance
audit cannot rule out overlap.

The simulation study is target- and score-informed, with no independent design
holdout. It covers two 3W event regimes, one univariate architecture, one
training budget, and one run per recipe. It compares full simulator recipes and
does not isolate oscillator dynamics from forcing-noise and augmentation
settings or test cross-channel modeling. Statistical alignment is assessed
using selected summary statistics, t-SNE projections, and window-level
nearest-neighbor checks with frequencies expressed in cycles per sample and
lags in samples, and the oscillator is
mechanism-inspired rather than physically calibrated. The observed
target-domain gains and GIFT-Eval loss describe specialization, not broad
superiority, deployment readiness, or regulatory compliance.

\clearpage
\section*{NeurIPS Paper Checklist}

\begin{enumerate}

\item {\bf Claims}
    \item[] Question: Do the main claims made in the abstract and introduction accurately reflect the paper's contributions and scope?
    \item[] Answer: \answerYes{}
    \item[] Justification: Section~\ref{sec:introduction} states the paper's
    contributions and limits its claims to the covered tasks, models, and 3W
    event regimes; Sections~\ref{sec:indeval} and~\ref{sec:simulation} report
    the corresponding evidence.
    \item[] Guidelines:
    \begin{itemize}
        \item The answer \answerNA{} means that the abstract and introduction do not include the claims made in the paper.
        \item The abstract and/or introduction should clearly state the claims made, including the contributions made in the paper and important assumptions and limitations. A \answerNo{} or \answerNA{} answer to this question will not be perceived well by the reviewers. 
        \item The claims made should match theoretical and experimental results, and reflect how much the results can be expected to generalize to other settings. 
        \item It is fine to include aspirational goals as motivation as long as it is clear that these goals are not attained by the paper. 
    \end{itemize}

\item {\bf Limitations}
    \item[] Question: Does the paper discuss the limitations of the work performed by the authors?
    \item[] Answer: \answerYes{}
    \item[] Justification: Limitations are stated in the main text alongside
    the claims they constrain. Section~\ref{sec:indeval} discloses IND-Eval's
    overlap with GIFT-Eval---three shared sources and substantial overlap in
    generic signal statistics---and reports that incomplete corpus disclosure
    prevents ruling out pretraining contamination. Findings are formulated
    within that evidence: the three-model comparison is illustrative,
    source-specific ranking reversals are treated as the relevant signal, and
    small metric gaps are not interpreted as meaningful differences.
    Section~\ref{sec:simulation} presents the 3W study as an exploratory,
    narrow case (two events, one architecture, one run per recipe, no
    independent design holdout) and states conclusions as a
    specialization--generality trade-off rather than broad superiority.
    Appendix~\ref{app:limitations} collects these limitations in more detail.
    \item[] Guidelines:
    \begin{itemize}
        \item The answer \answerNA{} means that the paper has no limitation while the answer \answerNo{} means that the paper has limitations, but those are not discussed in the paper. 
        \item The authors are encouraged to create a separate ``Limitations'' section in their paper.
        \item The paper should point out any strong assumptions and how robust the results are to violations of these assumptions (e.g., independence assumptions, noiseless settings, model well-specification, asymptotic approximations only holding locally). The authors should reflect on how these assumptions might be violated in practice and what the implications would be.
        \item The authors should reflect on the scope of the claims made, e.g., if the approach was only tested on a few datasets or with a few runs. In general, empirical results often depend on implicit assumptions, which should be articulated.
        \item The authors should reflect on the factors that influence the performance of the approach. For example, a facial recognition algorithm may perform poorly when image resolution is low or images are taken in low lighting. Or a speech-to-text system might not be used reliably to provide closed captions for online lectures because it fails to handle technical jargon.
        \item The authors should discuss the computational efficiency of the proposed algorithms and how they scale with dataset size.
        \item If applicable, the authors should discuss possible limitations of their approach to address problems of privacy and fairness.
        \item While the authors might fear that complete honesty about limitations might be used by reviewers as grounds for rejection, a worse outcome might be that reviewers discover limitations that aren't acknowledged in the paper. The authors should use their best judgment and recognize that individual actions in favor of transparency play an important role in developing norms that preserve the integrity of the community. Reviewers will be specifically instructed to not penalize honesty concerning limitations.
    \end{itemize}

\item {\bf Theory assumptions and proofs}
    \item[] Question: For each theoretical result, does the paper provide the full set of assumptions and a complete (and correct) proof?
    \item[] Answer: \answerNA{}
    \item[] Justification: The paper reports empirical benchmark and
    simulation results but makes no theoretical claims that require proofs.
    \item[] Guidelines:
    \begin{itemize}
        \item The answer \answerNA{} means that the paper does not include theoretical results. 
        \item All the theorems, formulas, and proofs in the paper should be numbered and cross-referenced.
        \item All assumptions should be clearly stated or referenced in the statement of any theorems.
        \item The proofs can either appear in the main paper or the supplemental material, but if they appear in the supplemental material, the authors are encouraged to provide a short proof sketch to provide intuition. 
        \item Inversely, any informal proof provided in the core of the paper should be complemented by formal proofs provided in appendix or supplemental material.
        \item Theorems and Lemmas that the proof relies upon should be properly referenced. 
    \end{itemize}

    \item {\bf Experimental result reproducibility}
    \item[] Question: Does the paper fully disclose all the information needed to reproduce the main experimental results of the paper to the extent that it affects the main claims and/or conclusions of the paper (regardless of whether the code and data are provided or not)?
    \item[] Answer: \answerYes{}
    \item[] Justification: The appendices document the benchmark construction,
    evaluation protocol, simulator, architecture, and training settings. The
    released Hugging Face artifact linked in Section~\ref{sec:introduction}
    provides the data, preprocessing scripts, environment locks, and execution
    instructions needed to reproduce the benchmark tasks. Exact checkpoint
    revisions were not retained, as disclosed in
    Appendix~\ref{app:inference}.
    \item[] Guidelines:
    \begin{itemize}
        \item The answer \answerNA{} means that the paper does not include experiments.
        \item If the paper includes experiments, a \answerNo{} answer to this question will not be perceived well by the reviewers: Making the paper reproducible is important, regardless of whether the code and data are provided or not.
        \item If the contribution is a dataset and\slash or model, the authors should describe the steps taken to make their results reproducible or verifiable. 
        \item Depending on the contribution, reproducibility can be accomplished in various ways. For example, if the contribution is a novel architecture, describing the architecture fully might suffice, or if the contribution is a specific model and empirical evaluation, it may be necessary to either make it possible for others to replicate the model with the same dataset, or provide access to the model. In general. releasing code and data is often one good way to accomplish this, but reproducibility can also be provided via detailed instructions for how to replicate the results, access to a hosted model (e.g., in the case of a large language model), releasing of a model checkpoint, or other means that are appropriate to the research performed.
        \item While NeurIPS does not require releasing code, the conference does require all submissions to provide some reasonable avenue for reproducibility, which may depend on the nature of the contribution. For example
        \begin{enumerate}
            \item If the contribution is primarily a new algorithm, the paper should make it clear how to reproduce that algorithm.
            \item If the contribution is primarily a new model architecture, the paper should describe the architecture clearly and fully.
            \item If the contribution is a new model (e.g., a large language model), then there should either be a way to access this model for reproducing the results or a way to reproduce the model (e.g., with an open-source dataset or instructions for how to construct the dataset).
            \item We recognize that reproducibility may be tricky in some cases, in which case authors are welcome to describe the particular way they provide for reproducibility. In the case of closed-source models, it may be that access to the model is limited in some way (e.g., to registered users), but it should be possible for other researchers to have some path to reproducing or verifying the results.
        \end{enumerate}
    \end{itemize}

\item {\bf Open access to data and code}
    \item[] Question: Does the paper provide open access to the data and code, with sufficient instructions to faithfully reproduce the main experimental results, as described in supplemental material?
    \item[] Answer: \answerYes{}
    \item[] Justification: The released Hugging Face artifact linked in
    Section~\ref{sec:introduction} provides the IND-Eval data and
    preprocessing code together with instructions for reproducing the
    benchmark tasks, while the appendices document the evaluation and
    simulation protocols.
    \item[] Guidelines:
    \begin{itemize}
        \item The answer \answerNA{} means that paper does not include experiments requiring code.
        \item Please see the NeurIPS code and data submission guidelines (\url{https://neurips.cc/public/guides/CodeSubmissionPolicy}) for more details.
        \item While we encourage the release of code and data, we understand that this might not be possible, so \answerNo{} is an acceptable answer. Papers cannot be rejected simply for not including code, unless this is central to the contribution (e.g., for a new open-source benchmark).
        \item The instructions should contain the exact command and environment needed to run to reproduce the results. See the NeurIPS code and data submission guidelines (\url{https://neurips.cc/public/guides/CodeSubmissionPolicy}) for more details.
        \item The authors should provide instructions on data access and preparation, including how to access the raw data, preprocessed data, intermediate data, and generated data, etc.
        \item The authors should provide scripts to reproduce all experimental results for the new proposed method and baselines. If only a subset of experiments are reproducible, they should state which ones are omitted from the script and why.
        \item At submission time, to preserve anonymity, the authors should release anonymized versions (if applicable).
        \item Providing as much information as possible in supplemental material (appended to the paper) is recommended, but including URLs to data and code is permitted.
    \end{itemize}

\item {\bf Experimental setting/details}
    \item[] Question: Does the paper specify all the training and test details (e.g., data splits, hyperparameters, how they were chosen, type of optimizer) necessary to understand the results?
    \item[] Answer: \answerYes{}
    \item[] Justification: Sections~\ref{sec:indeval}
    and~\ref{sec:simulation} summarize the protocols, while
    Appendices~\ref{app:ind-composition}--\ref{app:sim-arch} give task
    construction, metrics, checkpoints, architecture, optimization, and
    hyperparameters.
    \item[] Guidelines:
    \begin{itemize}
        \item The answer \answerNA{} means that the paper does not include experiments.
        \item The experimental setting should be presented in the core of the paper to a level of detail that is necessary to appreciate the results and make sense of them.
        \item The full details can be provided either with the code, in appendix, or as supplemental material.
    \end{itemize}

\item {\bf Experiment statistical significance}
    \item[] Question: Does the paper report error bars suitably and correctly defined or other appropriate information about the statistical significance of the experiments?
    \item[] Answer: \answerNo{}
    \item[] Justification: We report descriptive scores for fixed released
    checkpoints on the full IND-Eval benchmark, without confidence intervals
    or significance tests. The simulation study uses one training run per
    recipe; each run takes 24--48 hours on a single H100 or B200 GPU. These
    comparisons do not quantify variation across training seeds.
    \item[] Guidelines:
    \begin{itemize}
        \item The answer \answerNA{} means that the paper does not include experiments.
        \item The authors should answer \answerYes{} if the results are accompanied by error bars, confidence intervals, or statistical significance tests, at least for the experiments that support the main claims of the paper.
        \item The factors of variability that the error bars are capturing should be clearly stated (for example, train/test split, initialization, random drawing of some parameter, or overall run with given experimental conditions).
        \item The method for calculating the error bars should be explained (closed form formula, call to a library function, bootstrap, etc.)
        \item The assumptions made should be given (e.g., Normally distributed errors).
        \item It should be clear whether the error bar is the standard deviation or the standard error of the mean.
        \item It is OK to report 1-sigma error bars, but one should state it. The authors should preferably report a 2-sigma error bar than state that they have a 96\% CI, if the hypothesis of Normality of errors is not verified.
        \item For asymmetric distributions, the authors should be careful not to show in tables or figures symmetric error bars that would yield results that are out of range (e.g., negative error rates).
        \item If error bars are reported in tables or plots, the authors should explain in the text how they were calculated and reference the corresponding figures or tables in the text.
    \end{itemize}

\item {\bf Experiments compute resources}
    \item[] Question: For each experiment, does the paper provide sufficient information on the computer resources (type of compute workers, memory, time of execution) needed to reproduce the experiments?
    \item[] Answer: \answerYes{}
    \item[] Justification: Appendix~\ref{app:sim-arch} reports the optimizer
    steps, effective batch size, simulated-series count, GPU class, and
    24--48-hour wall-clock range for each recipe, and explains the
    simulator-throughput bottleneck behind that range.
    \item[] Guidelines:
    \begin{itemize}
        \item The answer \answerNA{} means that the paper does not include experiments.
        \item The paper should indicate the type of compute workers CPU or GPU, internal cluster, or cloud provider, including relevant memory and storage.
        \item The paper should provide the amount of compute required for each of the individual experimental runs as well as estimate the total compute. 
        \item The paper should disclose whether the full research project required more compute than the experiments reported in the paper (e.g., preliminary or failed experiments that didn't make it into the paper). 
    \end{itemize}
    
\item {\bf Code of ethics}
    \item[] Question: Does the research conducted in the paper conform, in every respect, with the NeurIPS Code of Ethics \url{https://neurips.cc/public/EthicsGuidelines}?
    \item[] Answer: \answerYes{}
    \item[] Justification: The authors reviewed the NeurIPS Code of Ethics and
    confirm that the research conforms to it. The work uses public numerical
    time-series sources and involves no human-subject research.
    \item[] Guidelines:
    \begin{itemize}
        \item The answer \answerNA{} means that the authors have not reviewed the NeurIPS Code of Ethics.
        \item If the authors answer \answerNo, they should explain the special circumstances that require a deviation from the Code of Ethics.
        \item The authors should make sure to preserve anonymity (e.g., if there is a special consideration due to laws or regulations in their jurisdiction).
    \end{itemize}

\item {\bf Broader impacts}
    \item[] Question: Does the paper discuss both potential positive societal impacts and negative societal impacts of the work performed?
    \item[] Answer: \answerYes{}
    \item[] Justification: Section~\ref{sec:introduction} discusses both.
    Positive impacts include reducing operational risk, cost, and unplanned
    downtime through more accurate industrial forecasts, and supporting the
    intended-purpose testing required for high-risk AI systems under the EU AI
    Act. Negative impacts follow from the same setting: aggregate leaderboards
    can mislead model selection for a particular process, and forecasting
    errors in industrial deployments can have safety consequences. The paper
    therefore argues that granular, source-specific evaluation is necessary,
    while stating that IND-Eval does not itself establish regulatory
    compliance.
    \item[] Guidelines:
    \begin{itemize}
        \item The answer \answerNA{} means that there is no societal impact of the work performed.
        \item If the authors answer \answerNA{} or \answerNo, they should explain why their work has no societal impact or why the paper does not address societal impact.
        \item Examples of negative societal impacts include potential malicious or unintended uses (e.g., disinformation, generating fake profiles, surveillance), fairness considerations (e.g., deployment of technologies that could make decisions that unfairly impact specific groups), privacy considerations, and security considerations.
        \item The conference expects that many papers will be foundational research and not tied to particular applications, let alone deployments. However, if there is a direct path to any negative applications, the authors should point it out. For example, it is legitimate to point out that an improvement in the quality of generative models could be used to generate Deepfakes for disinformation. On the other hand, it is not needed to point out that a generic algorithm for optimizing neural networks could enable people to train models that generate Deepfakes faster.
        \item The authors should consider possible harms that could arise when the technology is being used as intended and functioning correctly, harms that could arise when the technology is being used as intended but gives incorrect results, and harms following from (intentional or unintentional) misuse of the technology.
        \item If there are negative societal impacts, the authors could also discuss possible mitigation strategies (e.g., gated release of models, providing defenses in addition to attacks, mechanisms for monitoring misuse, mechanisms to monitor how a system learns from feedback over time, improving the efficiency and accessibility of ML).
    \end{itemize}
    
\item {\bf Safeguards}
    \item[] Question: Does the paper describe safeguards that have been put in place for responsible release of data or models that have a high risk for misuse (e.g., pre-trained language models, image generators, or scraped datasets)?
    \item[] Answer: \answerNA{}
    \item[] Justification: The work releases no high-risk generative model or
    scraped human-content dataset; it evaluates public numerical time-series
    sources.
    \item[] Guidelines:
    \begin{itemize}
        \item The answer \answerNA{} means that the paper poses no such risks.
        \item Released models that have a high risk for misuse or dual-use should be released with necessary safeguards to allow for controlled use of the model, for example by requiring that users adhere to usage guidelines or restrictions to access the model or implementing safety filters. 
        \item Datasets that have been scraped from the Internet could pose safety risks. The authors should describe how they avoided releasing unsafe images.
        \item We recognize that providing effective safeguards is challenging, and many papers do not require this, but we encourage authors to take this into account and make a best faith effort.
    \end{itemize}

\item {\bf Licenses for existing assets}
    \item[] Question: Are the creators or original owners of assets (e.g., code, data, models), used in the paper, properly credited and are the license and terms of use explicitly mentioned and properly respected?
    \item[] Answer: \answerYes{}
    \item[] Justification: The manuscript credits and cites the original
    dataset and model creators. The released Hugging Face artifact linked in
    Section~\ref{sec:introduction} includes per-source license files and
    transformation notices for every IND-Eval dataset;
    Chronos-2 and TiRex-2 are distributed under Apache-2.0 licenses documented
    on their model cards. TimesFM-3 uses the TimesFM Non-Commercial License
    v1.0 and is used here only for non-commercial research evaluation.
    \item[] Guidelines:
    \begin{itemize}
        \item The answer \answerNA{} means that the paper does not use existing assets.
        \item The authors should cite the original paper that produced the code package or dataset.
        \item The authors should state which version of the asset is used and, if possible, include a URL.
        \item The name of the license (e.g., CC-BY 4.0) should be included for each asset.
        \item For scraped data from a particular source (e.g., website), the copyright and terms of service of that source should be provided.
        \item If assets are released, the license, copyright information, and terms of use in the package should be provided. For popular datasets, \url{paperswithcode.com/datasets} has curated licenses for some datasets. Their licensing guide can help determine the license of a dataset.
        \item For existing datasets that are re-packaged, both the original license and the license of the derived asset (if it has changed) should be provided.
        \item If this information is not available online, the authors are encouraged to reach out to the asset's creators.
    \end{itemize}

\item {\bf New assets}
    \item[] Question: Are new assets introduced in the paper well documented and is the documentation provided alongside the assets?
    \item[] Answer: \answerYes{}
    \item[] Justification: The paper documents IND-Eval's construction,
    provenance, and limitations. The released Hugging Face artifact linked in
    Section~\ref{sec:introduction} provides a top-level benchmark README and
    per-source documentation, licenses, and preprocessing scripts alongside
    the released data.
    \item[] Guidelines:
    \begin{itemize}
        \item The answer \answerNA{} means that the paper does not release new assets.
        \item Researchers should communicate the details of the dataset\slash code\slash model as part of their submissions via structured templates. This includes details about training, license, limitations, etc. 
        \item The paper should discuss whether and how consent was obtained from people whose asset is used.
        \item At submission time, remember to anonymize your assets (if applicable). You can either create an anonymized URL or include an anonymized zip file.
    \end{itemize}

\item {\bf Crowdsourcing and research with human subjects}
    \item[] Question: For crowdsourcing experiments and research with human subjects, does the paper include the full text of instructions given to participants and screenshots, if applicable, as well as details about compensation (if any)? 
    \item[] Answer: \answerNA{}
    \item[] Justification: The work does not involve crowdsourcing or research
    with human subjects.
    \item[] Guidelines:
    \begin{itemize}
        \item The answer \answerNA{} means that the paper does not involve crowdsourcing nor research with human subjects.
        \item Including this information in the supplemental material is fine, but if the main contribution of the paper involves human subjects, then as much detail as possible should be included in the main paper. 
        \item According to the NeurIPS Code of Ethics, workers involved in data collection, curation, or other labor should be paid at least the minimum wage in the country of the data collector. 
    \end{itemize}

\item {\bf Institutional review board (IRB) approvals or equivalent for research with human subjects}
    \item[] Question: Does the paper describe potential risks incurred by study participants, whether such risks were disclosed to the subjects, and whether Institutional Review Board (IRB) approvals (or an equivalent approval/review based on the requirements of your country or institution) were obtained?
    \item[] Answer: \answerNA{}
    \item[] Justification: The work does not involve research with human
    subjects, so IRB or equivalent approval is not applicable.
    \item[] Guidelines:
    \begin{itemize}
        \item The answer \answerNA{} means that the paper does not involve crowdsourcing nor research with human subjects.
        \item Depending on the country in which research is conducted, IRB approval (or equivalent) may be required for any human subjects research. If you obtained IRB approval, you should clearly state this in the paper. 
        \item We recognize that the procedures for this may vary significantly between institutions and locations, and we expect authors to adhere to the NeurIPS Code of Ethics and the guidelines for their institution. 
        \item For initial submissions, do not include any information that would break anonymity (if applicable), such as the institution conducting the review.
    \end{itemize}

\item {\bf Declaration of LLM usage}
    \item[] Question: Does the paper describe the usage of LLMs if it is an important, original, or non-standard component of the core methods in this research? Note that if the LLM is used only for writing, editing, or formatting purposes and does \emph{not} impact the core methodology, scientific rigor, or originality of the research, declaration is not required.
    \item[] Answer: \answerNA{}
    \item[] Justification: The core methods consist of time-series evaluation,
    simulation, and forecasting; LLMs are not part of the research
    methodology.
    \item[] Guidelines:
    \begin{itemize}
        \item The answer \answerNA{} means that the core method development in this research does not involve LLMs as any important, original, or non-standard components.
        \item Please refer to our LLM policy in the NeurIPS handbook for what should or should not be described.
    \end{itemize}

\end{enumerate}

\end{document}